\pdfoutput=1

\documentclass[11pt, final]{article}

\usepackage[final]{acl}
\usepackage{booktabs}
\usepackage{times}
\usepackage{latexsym}

\usepackage[T1]{fontenc}

\usepackage[utf8]{inputenc}
\usepackage{booktabs}
\usepackage{colortbl}
\usepackage{color}
\usepackage{multirow}
\usepackage{pifont}
\usepackage{microtype}

\usepackage{inconsolata}

\usepackage{graphicx}
\usepackage{amsmath}
\usepackage{amssymb}
\usepackage{url}
\title{\textsc{PRAct}: Optimizing Principled Reasoning and Acting of LLM Agent}

\author{Zhiwei~Liu\thanks{zhiweiliu@salesforce.com}, Weiran~Yao\thanks{Equal contribution.}, Jianguo~Zhang, \text{Rithesh Murthy}, Liangwei~Yang, Zuxin~Liu, \\ \textbf{Tian~Lan, Ming~Zhu, Juntao~Tan, Shirley Kokane, Thai~Hoang, Juan~Carlos~Niebles,}  \\ \textbf{Shelby Heinecke,} \textbf{Huan Wang,} \textbf{Silvio Savarese}\and\textbf{Caiming Xiong} 
\\ {Salesforce AI Research, USA}}

\begin{document}
\maketitle
\begin{abstract}
We introduce the Principled Reasoning and Acting (PRAct) framework, a novel method for learning and enforcing action principles from trajectory data. Central to our approach is the use of text gradients from a reflection and optimization engine to derive these action principles. To adapt action principles to specific task requirements, we propose a new optimization framework, Reflective Principle Optimization (RPO). 
After execution, RPO employs a reflector to critique current action principles and an optimizer to update them accordingly.
We develop the RPO framework under two scenarios: Reward-RPO, which uses environmental rewards for reflection, and Self-RPO, which conducts self-reflection without external rewards. 
Additionally, two RPO methods, RPO-Traj and RPO-Batch, is introduced to adapt to different settings.
Experimental results across four environments demonstrate that the PRAct agent, leveraging the RPO framework, effectively learns and applies action principles to enhance performance. 
\end{abstract}

\section{Introduction}
Large language model (LLM) agents enable the action execution~\cite{autogpt23,goodman2023meta,yao2023react,wang2023survey} and consecutive reasoning ability~\cite{babyagi23,shinn2023reflexion,yao2023retroformer} of LLM.
Specifically, an LLM agent has both memory~\cite{shinn2023reflexion,li2023camel,liu2024agentlite} and action space~\cite{langchain23,wu2023autogen,liu2023bolaa}.  Adding those information into prompt extends the inference of LLM to be multi-turn action execution.
Therefore, an LLM agent is able to decide next actions based on its previous execution observations~\cite{wang2023drdt,xu2023rewoo,goodman2023meta,song2023llm}.

Optimizing the reasoning framework~\cite{yao2023react,liu2023bolaa,wang2023drdt} of agent is crucial in generating correct action execution.
As of now, customizing an LLM agent with existing open-source packages~\cite{liu2024agentlite,wu2023autogen,langchain23,Liu_LlamaIndex_2022} requires the designing of action spaces, such as function calls~\cite{patil2023gorilla} and code execution~\cite{wu2023autogen,wu2024stateflow}.
Along with a well-designed agent reasoning framework, i.e. the prompts of agent, an LLM is able to consecutively generate correct actions.
ReAct~\cite{yao2023react} framework achieves wide successes via adding one-step \textit{think} actions to enhance the reasoning ability of an agent. 
Additionally, Reflection~\cite{shinn2023reflexion,yao2023retroformer,paul2023refiner} mechanism is proposed to improve the agent self-correction capability.
\textit{Plan}~\cite{xu2023rewoo,liu2023bolaa} before execution is also verified to be beneficial.
\begin{figure}[t]
    \centering
    \includegraphics[width=1.0\linewidth]{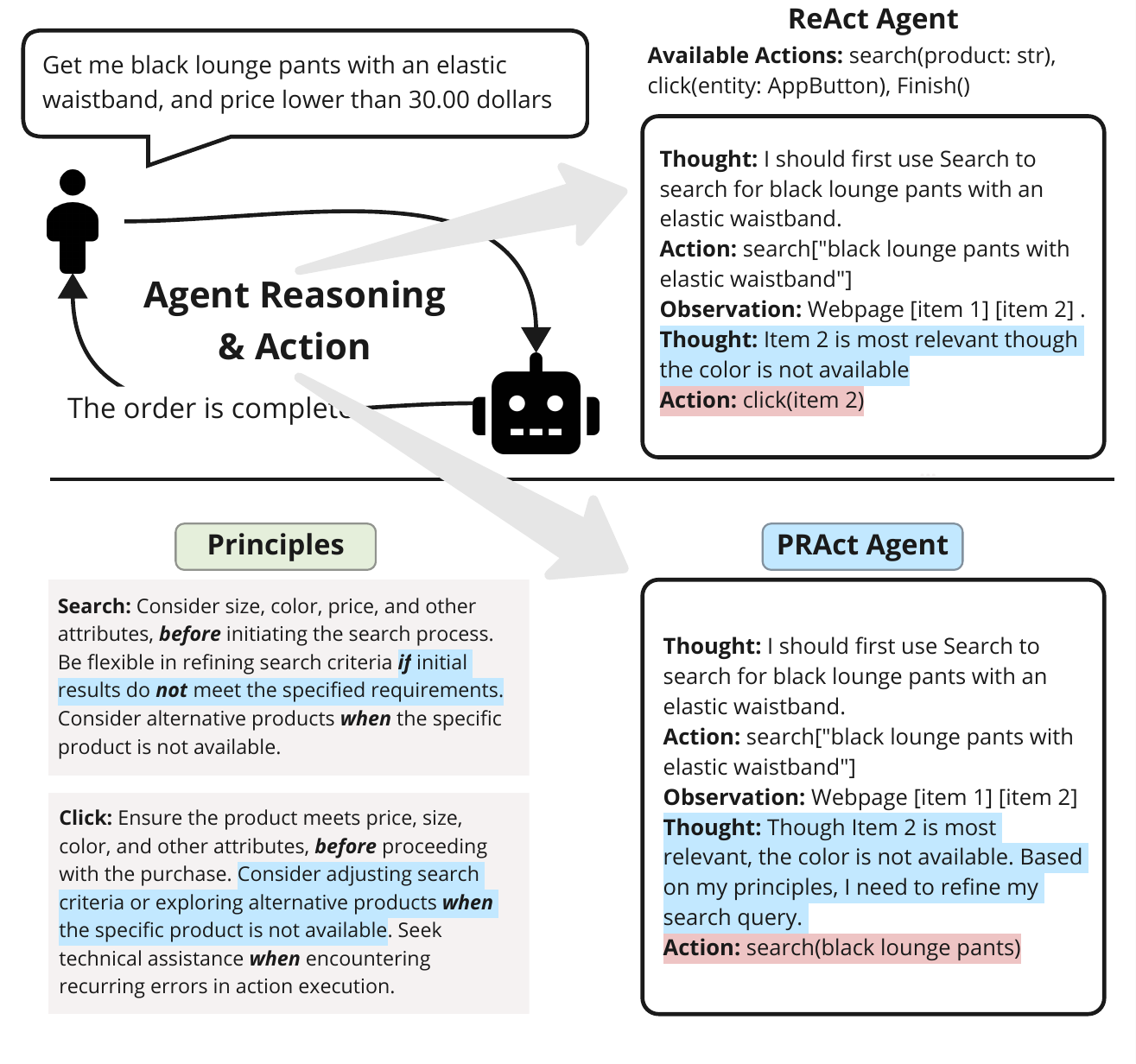}
    \caption{Comparison of ReAct and PRAct agents.}
    \label{fig:problem}
    \vspace{-0.5cm}
\end{figure}


Despite many successes, agent execution can fail to make decisions when faced with contradictory observations, particularly during the execution of long-step tasks. 
To address it, we propose a new type of reasoning strategy, \textit{PRAct}, for the LLM agent. Intuitively, we associate each action with principles that describe the conditions for using that action. 
During execution, an agent can check these principles before generating the next action. 
Compared to simple action descriptions, principles provide more detailed conditions on when to use the action and offer specific instructions on how to generate the parameters for an action.
We demonstrate the benefits of PRAct in Figure~\ref{fig:problem} via comparing with ReAct agent in WebShop~\cite{yao2022WebShop} where an agent uses search and click actions to interact with a shopping website.
The ReAct agent searches a query and, despite item 2 not having the available color, still clicks it as it appears most relevant. 
In contrast, the PRAct agent refines the search based on both search and click principles.
Consequently, the PRAct agent decides to search with an improved query, enhancing its decision-making process.

To reduce the labor involved in prompt design and to cover more complex scenarios, we propose a new principle optimization framework, Reflective Principle Optimization (RPO). RPO operates in three stages: execution, reflection, and optimization. 
During the execution stage, an agent performs tasks using predefined or null principles and memorizes the task trajectories. 
In the reflection stage, the agent reviews its task executions, evaluating how actions were selected and whether they met the task requirements. 
Finally, in the optimization stage, an optimizer refines principles to enhance agent performance. 
We investigate two optimization methods: RPO-Traj, which individually optimizes principles for each trajectory, and RPO-Batch, which concatenates all reflections in a batch for optimization. 


We summarize our contributions as follows: 1) PRAct is the first work that considers the action principles for LLM agent; 2) we propose two optimization methods to adapt the principles to tasks.

\section{\textsc{PRAct}: Optimizing Principled Reasoning and Acting}

\subsection{Formulation}
Given a task query, an agent is able to consecutively execute actions $[a_1, a_2, \dots, a_n]$ and collects observations $[o_1, o_2, \dots, o_n]$ from environments, where $o_i$ is the execution results of $a_i$. 
A policy function $\pi(a_t|c_t)$ predicts the next action $a_t$ given the execution trajectory context $c_t=[(a_1, o_1), (a_2, o_2), \dots, (a_{t-1}, o_{t-1})]$.
An Executor agent utilizes a language model to determine the policy function. 
It requires textual trajectory information for the prompt 
Intrinsically, those context information are text-based, including action names, action parameters and observations.

PRAct constraints the reasoning of LLM to follow a set of principles $\mathcal{P}$ as follows:
\begin{equation}
    \pi(a_t|c_t) = \text{Executor}(a_t|\mathcal{T}(c_t); \mathcal{P}),
\end{equation}
where $\mathcal{T}$ is the prompt template to organize context information and the principles $\mathcal{P}$ are guidelines that help shape the decision-making process of an LLM agent. 
Principles provide instructions on the usage of the action such as how to generate parameters for the action.
Additionally, principles reduce the set of potential actions by eliminating those that do not conform to the defined guidelines, thereby narrowing the search within the action space.
In this paper, we simplify the principles space to be the same as actions space, \textit{i.e.} each $a_{i}\in\mathcal{A}$ associated with a $p_i\in\mathcal{P}$.

\subsection{Reflective Principle Optimization~(RPO)}

\begin{figure*}[ht]
    \centering
    \includegraphics[width=0.9\textwidth]{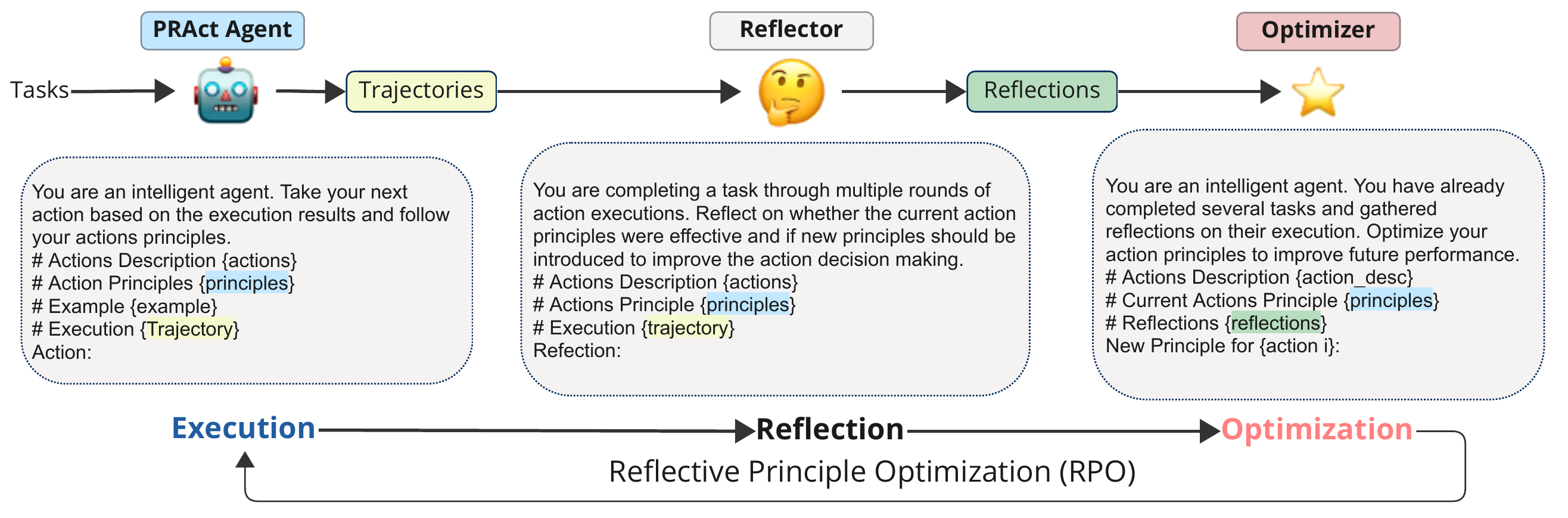}
    \caption{\textbf{PRAct and RPO overview.} Each iteration three stages: execution, reflection and optimization. During execution, an agent executes tasks with previous principles. The trajectories are saved.  Then, the agent reflects on those tasks executions. Finally, the agent leverages those self-reflection results to optimize the principle.}
    \label{fig:RPO}
    \vspace{-0.25cm}
\end{figure*}

Although the principles could be predetermined, as in the action descriptions, it is challenging to comprehensively cover all possible conditions without an automatic optimization paradigm. Therefore, we propose a new algorithm, Reflective Principle Optimization (RPO), to adapt principles for complex scenarios. RPO operates in three stages: 1) Execution, 2) Reflection, and 3) Optimization.

\subsubsection{Execution}
Given a set of tasks, the executor agent performs actions based on the current set of principles, collecting observations from the environment. 
This stage involves prompting the LLM agent to generate actions, which regressively calls Eq.~(2) until reaching the final actions or maximum steps. 
Given a task query $q$, we denote the trajectory as $c_{q}=[(a_q^{(1)}, o_q^{(1)}), (a_q^{(2)}, o_q^{(2)}), \cdot, (a_q^{(n)}, o_q^{(n)})]$.
Note that those actions may be some inner actions, such as \textit{think} or \textit{plan}~\cite{yao2023react,liu2024agentlite} , which do not forward to the environment and are associated with a default or null observation.
Executor collects a set of trajectory context sequences $\mathcal{C}$ for those queries $\mathcal{Q}$ during execution stage.

\begin{table*}[!t]
\centering
\caption{Overall comparison results. \textbf{Bold} denotes the best performance. \label{tab:results}}
\begin{tabular}{l|cccc|cccl}
\toprule
          & \multicolumn{4}{c|}{GPT-3.5-turbo}          & \multicolumn{4}{c}{GPT-4-turbo}             \\ \hline
& WebShop & Academia & Movie  & Weather & WebShop & Academia & Movie  & Weather \\ \hline
Act       & 0.4542  & 0.5304   & 0.5483 & 0.5869  & 0.5257  & 0.6704   & 0.5875 & 0.6882  \\
ReAct     & 0.4742  & 0.5504   & 0.5416 & 0.5973  & 0.5667  & 0.7428   & 0.5583 & 0.6990   \\
Reflexion & 0.5539  & 0.6024   & 0.5728 & 0.5876  & 0.5723  & 0.7796   & 0.6072 & 0.7197  \\
ExpeL & 0.5823	& 0.6318 &	0.6215	& 0.6475 & 0.6329	& 0.8084 &	0.6847	& 0.7583 \\
\rowcolor[HTML]{EFEFEF} 
\textbf{\texttt{PRAct-T}}   & \textbf{0.6012}  & 0.6798   & 0.6595 & 0.6953  & 0.6323  & \textbf{0.9207}   & 0.7132 & 0.7796  \\
\rowcolor[HTML]{EFEFEF} 
\textbf{\texttt{PRAct-B}}   & 0.5904  & \textbf{0.7396}   & \textbf{0.6625} & \textbf{0.7042}  & \textbf{0.6413}  & 0.8254   & \textbf{0.7250}  & \textbf{0.8331}  \\
\bottomrule
\end{tabular}
\end{table*}

\subsubsection{Self-Reflection}
After executing the actions, a reflector agent reflects on trajectories $\mathcal{C}$ by analyzing the collected observations.
This stage involves evaluating the effectiveness of the actions in each trajectory and the adherence to the principles as follows:
\begin{equation}
    r_q = \textsc{Reflector}(c_q, \mathcal{P}),
\end{equation}
for all $c_q\in\mathcal{C}$.
The reflection process identifies conditions or guidelines where the principles need adjustment to better handle the observed tasks.
If an environment provides rewards toward the execution,
it is a reward-based reflector aligning the executions with reward feedback. 
Instead, if no rewards present for execution, it is a self-reflector.

\subsubsection{Optimization}
Based on the reflection results, we leverage the generation ability of LLM to refine the principles for improving the performance of agent in similar future scenarios.
This stage involves refining the principles to better align with the observed conditions and enhance decision-making. 
We investigate two types of optimization methods.\\
\textbf{RPO-Traj}. This approach individually considers each trajectory and its reflection to optimize principles. 
Then a batch of principles are summarized as a new set of tailored principles $\mathcal{P}^{*}$.
We formulate RPO-Traj as follows:
\begin{equation}
    \mathcal{P}^{*}=\sum_{\mathcal{Q}}\textsc{OPT}(r_q, \mathcal{P}),
\end{equation}
where $\sum_{\mathcal{Q}}$ denotes a summarizor of all principles generated from optimizer OPT for all queries $\mathcal{Q}$. \\
\textbf{RPO-Batch}. We use a prompt template to concatenate all the reflections in a batch.
Then the optimizer directly generates new principles via considering all those reflections, which is formulated as follows: 
\begin{equation}
    \mathcal{P}^{*}=\textsc{OPT}(\textsc{CONCAT}{\{r_q | q\in \mathcal{Q}\}}, \mathcal{P}),
\end{equation}
where $\textsc{CONCAT}$ denotes using a prompt template to concat those reflections. 
In comparison, RPO{-Traj} requires generating principles for $|Q|+1$ times, while RPO{-Batch} only needs one time principles generation but with $|Q|$ times longer context length. 
Hence, long context reasoning ability is necessary for an optimizer in RPO{-Batch} method. 

\section{Experiment}

\subsection{Experiment Setup}
\textbf{Baselines}. We compare our PRAct agent with existing Act, ReAct~\cite{yao2023react}, Reflexion~\cite{shinn2023reflexion} agent reasoning methods and Expel~\cite{zhao2024expel} prompt optimization framework. 
In this paper, we employ GPT-3.5-Turbo-0125 and GPT-4-Turbo-2024-04-09~\cite{openai2023gpt4}
as two foundation LLMs. And for simplicity, 
the executor, reflector and optimizer in PRAct are of the same language model.
\\
\textbf{Benchmarks and Evaluation}. Following AgentBoard~\cite{ma2024agentboard}, we evaluate PRAct agent on three tool environments and one WebShop environment. Tool environments support the designing of \textsc{Weather}, \textsc{Movie}, and \textsc{Academia} agents. Tasks are 60 queries and actions are a set of function calls. 
The reward score is the recall of ground truth actions.
Webshop environment is a web browser simulation. Agent performs either \textit{search} and \textit{click} actions to complete 251 online shopping tasks.
Reward is attributes coverage ratio between final shopped items and ground truth item. 
\subsection{Optimization setup}
For optimizing the WebShop agent with a Reward-based reflector, we randomly split the query tasks into training, validation, and test tasks with a ratio of 3:1:1. 
During each training step, we sample a batch of training tasks to execute and use RPO to optimize the principles. Performance on validation tasks is used for early stopping, and results are reported on test tasks.
For tool agents, we use a self-reflector without rewards, making reflection tasks the same as test tasks. 
Since there is no ground truth, no data leakage problem exists. 
We tune the training batch size in [10,20,40] for WebShop and [2,4,6] for tool environments.

\subsection{Experiment Results}

\paragraph{Overall Performance.}

We present comprehensive comparisons of our methods against the agent baselines in Table~\ref{tab:results}. 
PRAct-T and PRAct-B are our methods with RPO-Traj and RPO-Batch optimization methods, respectively. 
We observe consistently better performance of PRAct agent, which demonstrates the effectiveness of principles in improving agent performance. 
Between the two optimization methods, \textit{i.e.} PRAct-T and PRAct-B, PRAct-B generally performs better than PRAct-T. The reason is that summarizing principles from a batch of reflections enables potential reasoning across trajectories. 
However, PRAct-T outperforms PRAct-B due to the potential weaker long context understanding ability of GPT-3.5-Turbo, which indicates batch-wise optimization is more suitable for larger models. 
\begin{table}[]
    \centering
    \begin{tabular}{l|c|c}
        \toprule
        Reflector & GPT-3.5 & GPT-4 \\
        \midrule
        Self-T & 0.5871 &  0.6172 \\
        Self-B & 0.5763 &  0.6238 \\
        Reward-T & \textbf{0.6012} & 0.6323 \\
        Reward-B & 0.5904 & \textbf{0.6413} \\
        \bottomrule
    \end{tabular}
    \caption{Different reflectors of PRAct. \textit{Self} and \textit{Reward} stand for self and reward-based reflectors, respectively. T and B denote RPO-Traj and RPO-Batch, respectively.}
    \label{tab:reflectors}
\end{table}

An additional variant is \textit{PRAct with self-reflector} on Webshop. We compare it on both RPO-T and RPO-B optimization methods, and report the results in Table~\ref{tab:reflectors}.
Compared with both results, reward-based reflector, demonstrates its superiority in optimizing principles with rewards. 

\paragraph{Optimization Curve.}

\begin{figure}[h]
    \centering
    \includegraphics[width=\linewidth]{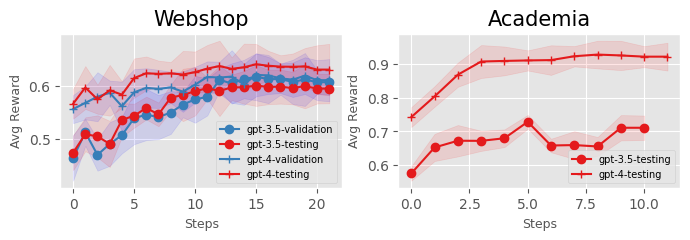}
    \caption{Training curves in Webshop and Academia with different LLMs and data splits. The reported scores are the average across 5 random seeds.}
    \label{fig:curve}
\end{figure}

We present the training curves in Fig.~\ref{fig:curve}. Although at each step, we did not pick the best principle out of the sampled action principles on the validation set,  we still observe consistent improvement over time. Notably,  with action principles optimized by PRAct, LLM agents under GPT-3.5-Turbo can match the performance of GPT-4-turbo in Webshop environment. 

\section{Conclusion}
We propose a novel agent reasoning framework, PRAct, which provides principles of actions and thus benefits the action understanding of agent.
Besides, we introduce two optimization algorithm, RPO-Traj and RPO-Batch for adapting the action principles with task executions. 
Experimental results on four environments demonstrates the effectiveness of PRAct framework. 
And the training curve illustrates the learning efficacy of RPO. 
In conclusion, PRAct opens a new discussion on how to regularize the agent actions while RPO shades the light on how to optimize the agent prompts. 

\bibliography{citation}

\clearpage
\end{document}